\documentclass{article} 
\usepackage{iclr2026_conference,times}

\usepackage{amsmath,amsfonts,bm}

\def\eqref#1{equation~\ref{#1}}

\def\1{\bm{1}}

\DeclareMathAlphabet{\mathsfit}{\encodingdefault}{\sfdefault}{m}{sl}
\SetMathAlphabet{\mathsfit}{bold}{\encodingdefault}{\sfdefault}{bx}{n}

\usepackage{graphicx}
\usepackage{hyperref}
\usepackage{url}
\usepackage{amssymb}
\usepackage{algorithm}
\usepackage{algpseudocode}
\usepackage{multirow}
\usepackage{wrapfig}
\usepackage{caption}
\usepackage{booktabs} 
\usepackage{subcaption}
\usepackage{makecell} 
\usepackage{colortbl,xcolor}
\usepackage{pifont}
\definecolor{lightblue}{rgb}{0.9,0.95,1}
\title{AdaNav: Adaptive Reasoning with Uncertainty for Vision-Language Navigation}

\newcommand{\mone}{\ding{172}} 
\newcommand{\mtwo}{\ding{173}} 
\newcommand{\mthree}{\ding{174}} 
\newcommand{\mfour}{\ding{175}} 
\newcommand{\sys}[0]{AdaNav}

\author{
  \fontsize{9pt}{5pt}\selectfont
  Xin Ding\textsuperscript{1}\thanks{xinding64@mail.ustc.edu.cn}
  \quad
  Jianyu Wei\textsuperscript{1}
  \quad
  Yifan Yang\textsuperscript{2}
  \quad
  Shiqi Jiang\textsuperscript{2}
  \quad
  Qianxi Zhang\textsuperscript{2}
  \quad 
  Hao Wu\textsuperscript{3}
  \quad 
  Fucheng Jia\textsuperscript{4,6}
  \quad \\
  Liang Mi\textsuperscript{6}
  \quad
  Yuxuan Yan\textsuperscript{5}
  \quad 
  Weijun Wang\textsuperscript{6}
  \quad
  Yunxin Liu\textsuperscript{6}
  \quad
  Zhibo Chen\textsuperscript{1} \textsuperscript{$\dag$}
  \quad
  Ting Cao\textsuperscript{6} \textsuperscript{$\dag$} \\
  \textsuperscript{1}University of Science and Technology of China
    \quad
  \textsuperscript{2} Microsoft Research
  \quad 
  \textsuperscript{3}Nanjing University
  \quad \\
  \quad\quad\quad\quad\quad\quad\quad\quad\quad\textsuperscript{4} Central South University
  \quad 
  \textsuperscript{5}\mbox{Zhejiang University}
  \quad \\
  \quad\quad\quad\quad\quad\quad\textsuperscript{6}\mbox{Institute for AI Industry Research (AIR), Tsinghua University}
}

\newcommand{\baseline}[1]{\cellcolor{lightblue}{#1}}

\iclrfinalcopy 
\begin{document}

\maketitle
\begingroup
\renewcommand\thefootnote{}
\footnotetext{$^{\dag}$Corresponding Author. tingcao@mail.tsinghua.edu.cn}
\addtocounter{footnote}{-1}
\endgroup

\begin{abstract}
Vision-Language Navigation (VLN) requires agents to follow natural language instructions by grounding them in sequential visual observations over long horizons. Explicit reasoning could enhance temporal consistency and perception–action alignment, but reasoning at fixed steps often leads to suboptimal performance and unnecessary computation. To address this, we propose \textbf{AdaNav}, an uncertainty-based adaptive reasoning framework for VLN. At its core is the \textbf{Uncertainty-Adaptive Reasoning Block} (UAR), a lightweight plugin that dynamically triggers reasoning. We introduce \textit{Action Entropy} as a policy prior for UAR and progressively refine it through a \textit{Heuristics-to-RL} training method, enabling agents to learn difficulty-aware reasoning policies under the strict data limitations of embodied tasks. Results show that with only \textit{6K} training samples, AdaNav achieves substantial gains over closed-source models trained on \textit{million-scale} data, improving success rate by 20\% on R2R val-unseen, 11.7\% on RxR-CE, and 11.4\% in real-world scenes. The code is available at \href{https://github.com/xinding-sys/AdaNav}{AdaNav}.


\end{abstract}

\section{Introduction}

As a fundamental capability for embodied agents, Vision-Language Navigation (VLN) requires agents to interpret natural language instructions and continuously ground them in sequential visual observations to execute long-horizon navigation trajectories~\citep{gu2022vision,park2023visual}. 
Existing VLM-based methods either rely on augmenting navigation with auxiliary modalities~\citep{krantz2021waypoint,xu2023vision,yin2024sg}, such as depth, odometry, or topological maps to strengthen spatial understanding, or instead scale up training on VLN data \textit{without} auxiliary inputs to improve quality and  generalization~\citep{zheng2024towards,wei2025streamvln,yu2025correctnav}. However, despite these advances, current methods still hindered by two major challenges of VLN: (1) Consistent temporal grounding: continuously capturing progress along the trajectory, tracking completed parts, and deciding the next action; (2) Robust perception–action mapping: grounding language in the current spatial context, recognizing landmarks, localizing itself, and selecting appropriate navigation actions.

To address these challenges, explicit reasoning has been introduced to VLN~\citep{zhou2024navgpt,wang2024llm,lin2025navcot,chen2024mapgpt}, enabling agents to better align language, perception, and action over long-horizon navigation trajectories. However, current straightforward reasoning at each step not only incurs prohibitive computational overhead, but also results in overthinking~\citep{sui2025stop,wu2025more,shen2025dast} that degrades navigation quality (Figure~\ref{fig:reasoning_analysis} and Table~\ref{table:different_data_performance} show higher quality with fewer reasoning steps).  
Ideally, VLN agents should exhibit adaptive reasoning, i.e., deciding \textit{when and how} to reason. However, achieving such adaptivity and mitigating the overconfidence issue of LLMs~\citep{sun2025large,groot2024overconfidence,yoo2024much} typically require large-scale supervised fine-tuning (SFT) with task-specific data~\citep{wen2024mitigating,lin2025onetwovla}. However, embodied interaction data is costly to collect and far from web-scale. Under such limited data conditions, it remains difficult for models to learn when and how to adaptively invoke reasoning.

\begin{figure}[t]
\centering
\includegraphics[width=\textwidth]{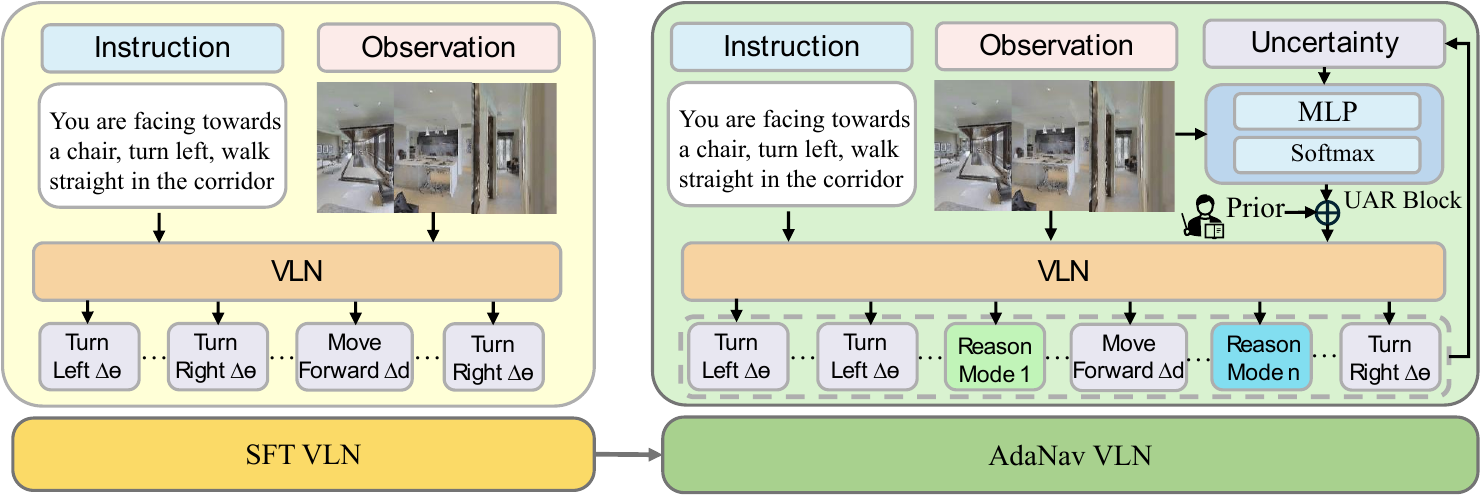}
\caption{AdaNav augments a base VLN model by integrating the Uncertainty-Adaptive Reasoning Block (UAR Block). UAR Block leverages model uncertainty to autonomously trigger reasoning modes and timing, enhancing consistent temporal grounding and perception–action mapping while significantly improving efficiency and mitigating overthinking.}
\label{fig:overview}
\end{figure}
\begin{figure}[t]
\centering
\includegraphics[width=\textwidth]{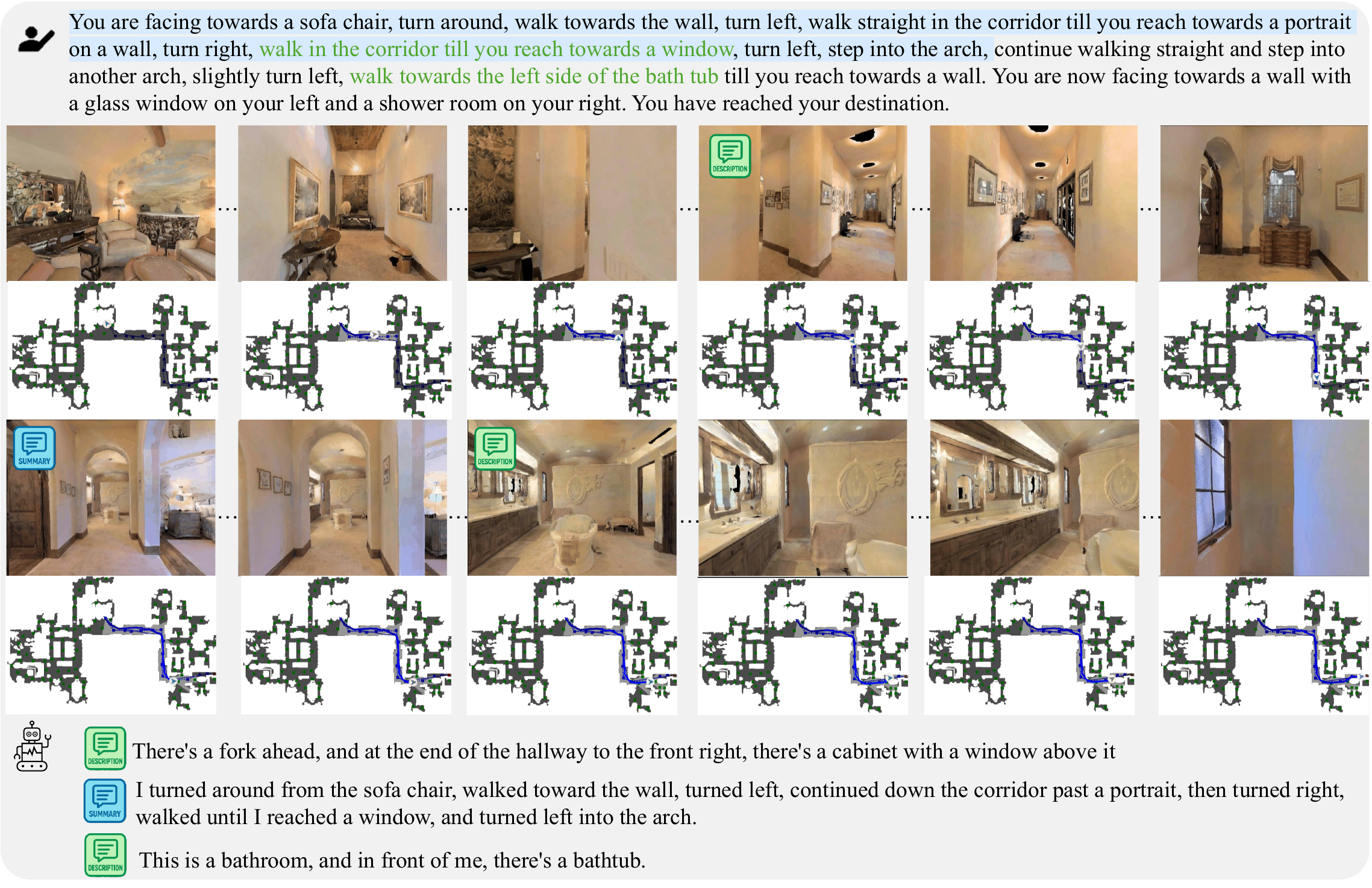}
\vspace{-2mm}
\caption{A visualization example of \sys{}. It autonomously invokes reasoning, e.g., summarization and description when necessary to maintain consistent temporal grounding and robust perception–action mapping.}
\label{fig:visual_reason_example}
\vspace{-6mm}
\end{figure}

To avoid the data limitation, we propose \textbf{uncertainty-based adaptive reasoning for navigation (AdaNav)}, as shown in Figure~\ref{fig:overview}. By defining \textit{Action Entropy} as an indicator for uncertainty, \sys{} utilizes this as an objective and interpretable heuristic prior to decide when and how to reason, and then refine this prior gradually through reinforcement learning (RL) to optimize the reasoning trigger policy. By combining the efficiency of heuristic guidance with the optimality of RL, \sys{} do not involve costly labeled reasoning triggering data, but enable the agent to automatically invoke reasoning when necessary to maintain temporal grounding and robust perception–action mapping in the long-horizon navigation. See Figure~\ref{fig:visual_reason_example} as an example. 

To realize \sys{}, we introduce a \textit{Uncertainty-Adaptive Reasoning Block (UAR Block)} and the \textit{Heuristic-to-RL} training mechanism. UAR block, as a plugin for available VLN models, collects historical, embodied-interaction-dependent uncertainty signals and generates vectorized control signals to dynamically trigger VLN for appropriate reasoning modes. Leveraging the interpretable signals from the UAR Block, the Heuristic-to-RL training first explores the action space under these heuristic priors (e.g., triggering reasoning when uncertainty exceeds a threshold) to guide decision-making at critical moments. As training progresses, the influence of these priors is gradually annealed, allowing RL to autonomously refine the UAR reasoning-trigger policy for optimal reward.


To demonstrate the effectiveness of AdaNav, we integrate it with state-of-the-art open-source VLN backbones and evaluate on classic benchmarks. Remarkably, \textbf{with only 6K training samples, AdaNav significantly surpasses closed-source models trained on million-scale data.} Specifically, our method achieves an average 20\% improvement in success rate on R2R val-unseen~\citep{krantz2020beyond}, and even without training on the larger and more challenging RxR-CE~\citep{ku2020room}, \sys{} yields a 11.7\% gain, demonstrating the cross-dataset generality. Additionally, AdaNav exhibits strong robustness in Sim-to-Real deployment, achieving approximately a 11.4\% success rate improvement over 150 instructions across four \textbf{real-world indoor scenes}. As training proceeds, \sys{} reduces the average number of reasoning steps per trajectory to only \textbf{2.5} 
(over trajectories with an average length of \textbf{55 steps}), while the success rate increases 7\% compared to reasoning at fixed steps. 
Notably, 71\% of reasoning steps are concentrated on hard trajectories. These results indicate that training makes reasoning more difficulty-aware and mode-adaptive.

\section{Related Work}
\paragraph{VLN with Auxiliaries.}
VLN requires agents to follow free-form linguistic instructions and visual cues to reach a target location. Early studies relied on pre-defined waypoints for discrete navigation~\citep{qi2020reverie,thomason2020vision} in the Habitat-Matterport3D simulator~\citep{chang2017matterport3d}, while more recent works~\citep{qi2020object,an2021neighbor,hong2020language,tan2019learning,wang2019reinforced} use continuous environments, namely \textit{VLN-CE}, like Habitat~\citep{krantz2020beyond}, enabling low-level actions (e.g., move forward, rotate) for more realistic navigation.  
With the rise of Transformers, many works introduced pre-trained methods with auxiliary modalities, e.g., depth, odometry, or topological maps, for VLN~\citep{ma2019self,wang2019reinforced}. DUET~\citep{chen2022think} and ETPNav~\citep{an2024etpnav} build topological maps for global navigation understanding, while GridMM~\citep{wang2023gridmm} introduced a dynamic egocentric grid memory. Although these methods improve spatial awareness, they inevitably limit generalization and introduce computational overhead and noise~\citep{zhang2024navid}. Modern works increasingly target video-only general solutions for VLN without auxiliaries~\citep{zhang2024navid,cheng2024navila,zhang2024uni}. \textit{VLN-CE with only videos captured by the monocular camera is also the target of this paper.}   

\vspace{-2mm}
\paragraph{Vision-Language Models for Navigation.} 
With the rapid development of Vision-Language Models (VLMs), RT-2~\citep{zitkovich2023rt} demonstrates the potential of transferring web-scale knowledge from VLMs to generalizable robotic manipulation. Recent work has focused on scaling VLN training data and fine-tuning large VLMs. For example, Navid~\citep{zhang2024navid} used 550k navigation samples to fine-tune Vicuna for navigation; NaVILA~\citep{cheng2024navila} expanded to 3–5M samples combining real and simulated navigation data plus general VQA supervision; Uni-NaVid~\citep{zhang2024uni} further incorporated 3.6M multi-task trajectories from Habitat-Matterport3D~\citep{chang2017matterport3d,krantz2020beyond} and real video QA data~\citep{azuma2022scanqa,chen2024panda,li2024llama} for cross-task generalization. Despite these advances, VLM-based VLN agents still fall short in task quality, struggling with consistent temporal grounding and robust perception–action mapping, particularly in long-horizon trajectories and complex environments.

\vspace{-2mm}
\paragraph{Explicit Reasoning for Navigation.}


To mitigate these challenges, recent works introduce explicit reasoning via off-the-shelf LLMs, where pre-defined programming rules constrain when and how reasoning modes—description, summarization, or error correction—are applied. For example, LLM-Planner~\citep{song2023llm} parses instructions into sub-goals; NavGPT~\citep{zhou2024navgpt} generates step-wise textual scene descriptions and historical trajectories; NavGPT-2~\citep{zhou2024navgpt2} further integrates visual grounding; MiC~\citep{qiao2023march} organizes reasoning into a “summarization–planning–correction” loop; DiscussNav~\citep{long2024discuss}, MCGPT~\citep{zhan2024mc}, and InstructNav~\citep{long2024instructnav} leverage expert collaboration or memory graphs for error correction and historical summarization.

While these rule-driven frameworks offer interpretability, they inherently restrict flexibility in open-ended environments, hinder efficiency, and may lead to overthinking~\citep{fang2025thinkless,dai2025s}. In contrast, our method will employ a learnable mechanism that enables agents to autonomously decide when and how to reason. 

\section{Method}

\subsection{Problem Formulation of AdaNav}

The central problem investigated in this work is how to enable an embodied agent to adaptively decide \emph{when} and \emph{how} to invoke reasoning during VLN. Unlike conventional approaches that either disable reasoning or enforce rule-based reasoning at fixed steps, our goal is to learn an autonomous reasoning policy that dynamically determines the timing and mode of reasoning, optimizing both efficiency and navigation performance.

\paragraph{Vision-Language Navigation.}  
We consider a standard VLN setting where an agent is placed in a 3D environment $\mathcal{E}$ with state space $\mathcal{S}$ and action space 
$\mathcal{A} = \{\text{turn\_left}(\Delta \theta), \text{turn\_right}(\Delta \theta), \text{move\_forward}(\Delta d), \text{stop}\}$, 
where $\Delta \theta$ and $\Delta d$ denote the angle and distance, respectively. Given a natural language instruction $I$ and sequential visual observations $\{o_1, o_2, \dots\}$, the agent executes a trajectory $\tau=\{(s_t,a_t)\}_{t=1}^H$ toward a goal $s^*$ specified implicitly by $I$, aiming to maximize task success:
\begin{equation}
\pi^* = \arg\max_{\pi}\; \mathbb{E}_{\tau \sim \pi} \big[ \mathbf{1}(s_H = s^*) \big] .
\end{equation}
\paragraph{Adaptive Reasoning Navigation.}
To improve VLN performance in long-horizon and complex environments, we allow explicit reasoning at step $t$ with a \emph{mode} variable $m_t \in \{\varnothing\}\cup\mathcal{M}$ and reasoning content $r_t$. Here, $m_t=\varnothing$ denotes \emph{no reasoning} (so $r_t=\varnothing$), while $m_t\in\mathcal{M}$ denotes invoking a reasoning mode from a predefined set. In this work, we consider three reasoning modes: \emph{description}, \emph{summary}, and \emph{error correction} (see Figure~\ref{fig:visual_reason} and Appendix~\ref{app:visual section appendix} for instances). The agent’s policy then consists of two coupled processes:  
1) a navigation policy $\pi_{\text{nav}}(a_t \mid h_t^{\text{nav}}, I, r_{\le t})$, and  
2) a reasoning policy $\pi_{\text{rea}}(m_t, r_t \mid h_t^{\text{rea}}, I)$ that jointly decides \emph{when} to reason (via $m_t=\varnothing$ vs.\ $m_t\neq\varnothing$) and \emph{which mode} to use (via $m_t\in\mathcal{M}$).

The overall joint policy is
\begin{equation}
\pi^*(a_t, m_t, r_t \mid h_t, I)
= \pi_{\text{nav}}(a_t \mid h_t^{\text{nav}}, I, r_{\le t}) \cdot \pi_{\text{rea}}(m_t, r_t \mid h_t^{\text{rea}}, I)
\end{equation}
where $h_t = (h_t^{\text{nav}}, h_t^{\text{rea}})$ denotes the full history, with $h_t^{\text{nav}}$ and $h_t^{\text{rea}}$ representing the navigation-related and reasoning-related information, respectively.
For clarity, we factorize the reasoning policy as:
\begin{equation}
\label{equ:ada_reason}
\pi_{\text{rea}}(m_t, r_t \mid h_t^{\text{rea}}, I)
= \underbrace{\pi_{\text{txt}}(r_t \mid m_t, h_t^{\text{rea}}, I)}_{\text{reasoning content}}
\cdot
\underbrace{\pi_{\text{sel}}(m_t \mid h_t^{\text{rea}}, I)}_{\text{when/which mode}}
\end{equation}
With the constraint $r_t=\varnothing$ if $m_t=\varnothing$. Here, $\pi_{\text{txt}}$ shares the same network as $\pi_{\text{nav}}$.

By integrating navigation and reasoning, the overall learning objective is to jointly optimize both policies, aiming to maximize task performance while maintaining computational efficiency.
\begin{equation}
\label{equ:adaptive_reason}
\pi^* = \arg\max_{(\pi_{\text{rea}}, \pi_{\text{nav}})} \;
\mathbb{E}_{\tau \sim (\pi_{\text{nav}}, \pi_{\text{rea}})}
\Big[ R_{\text{task}}(\tau) \Big]
\end{equation}
where $R_{\text{task}}(\tau)$ jointly accounts for navigation success (e.g., progress or success indicator) and the latency penalty induced by reasoning calls.

\subsection{Methodology of AdaNav}
\paragraph{Motivation.}
Adaptive reasoning requires the agent to selectively decide \emph{when} reasoning is beneficial and \emph{which mode} to invoke. However, native VLMs are neither sensitive nor objective in perceiving task difficulty, often resulting in overconfidence. In LLM research, similar issues (e.g., in mathematical reasoning) have been alleviated by incorporating high-quality reasoning traces with supervised fine-tuning~\citep{zhang2025continue,tian2025think,guo2025deepseek}. In contrast, for embodied agents, collecting such high-quality interaction traces is prohibitively expensive. This motivates the development of alternative approaches that enable agents to acquire adaptive reasoning capabilities without relying on large-scale reasoning supervision.

To this end, we propose Adaptive Reasoning Navigation (AdaNav), which leverages interpretable uncertainty signals to dynamically trigger reasoning only when necessary. By combining a learnable reasoning policy with a navigation policy, AdaNav enables efficient, difficulty-aware, and mode-adaptive reasoning, achieving high performance in long-horizon and complex VLN tasks.

\subsubsection{Uncertainty-Adaptive Reasoning Block}




Recent works~\citep{kazemnejad2024vineppo,wang2025beyond,fu2025deep} in language reasoning have shown that high-entropy tokens exert a disproportionately large influence on single-step text generation. Inspired by this, we explore whether a similar principle can serve as a signal for identifying “forking steps” in navigation. Specifically, we define action entropy $H$ as an uncertainty measure: 
\begin{equation}
H = - \frac{1}{N} \sum_{t=1}^{N} \sum_{v=1}^{V} p_t[v] \log p_t[v],
\end{equation}
where $N$ is the number of tokens generated, $V$ is the size of the vocabulary, and $p_t[v]$ is the probability of the $v$-th vocabulary token at time step $t$.

\begin{figure}[t]
\centering
\includegraphics[width=0.9\textwidth]{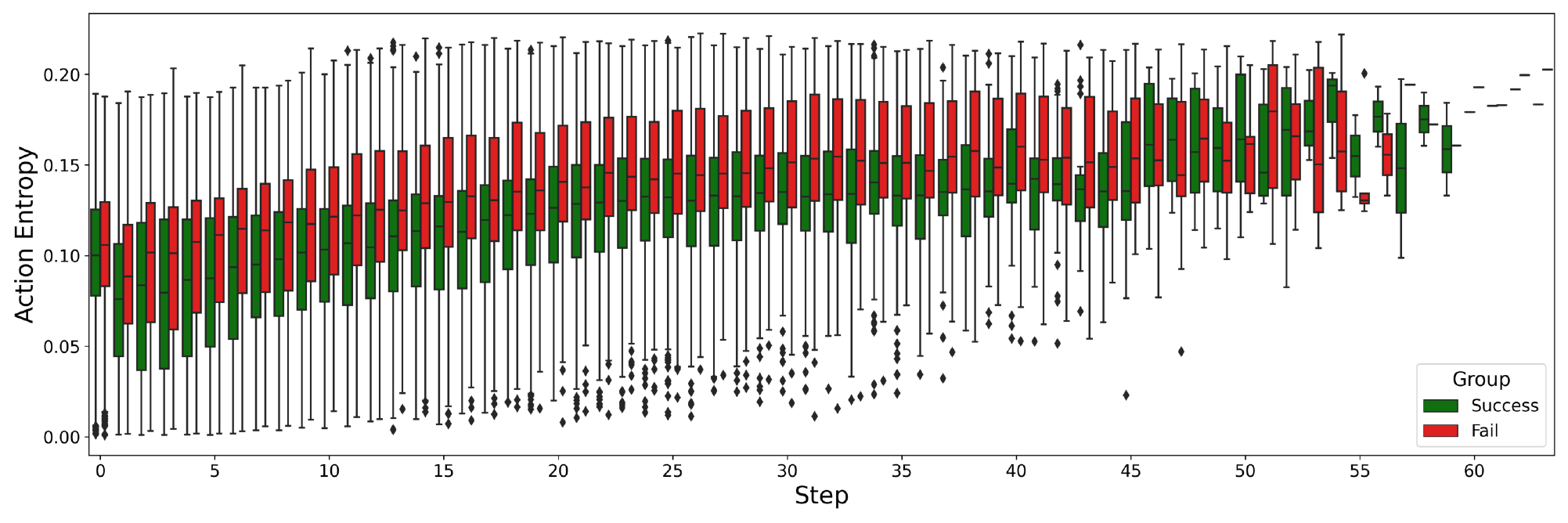}
\vspace{-4mm}
\caption{Action entropy per step for successful (green) vs. failed (red) trajectories. Failed trajectories show higher, especially in later steps, indicating that policy uncertainty correlates with navigation errors.}
\label{fig:entropy_step}
\vspace{-5mm}
\end{figure}

To validate the effectiveness of action entropy, we conduct a diagnostic study on navigation trajectories. As shown in Figure~\ref{fig:entropy_step}, episodes with high and sustained entropy are strongly correlated with failures, while successful trajectories maintain consistently low entropy (Different Means). Furthermore, instantaneous entropy alone is insufficient: short-lived spikes do not necessarily imply failure, and many successful trajectories exhibit temporary fluctuations without requiring reasoning (Comparable Extremes). 

Conversely, \textit{combining historical action entropy trends with current action entropy states provides a more reliable signal $H_{\le t}$}: successful trajectories show relatively lower entropy over time, while failure-prone ones accumulate persistently high entropy. 


\paragraph{Method Design.}
Motivated by these findings, we design a lightweight Uncertainty-Adaptive Reasoning Block that fuses $H_{\le t}$ with the current observation $o_t$, forming the reason-related information $h_t^{\text{rea}} = (H_{\le t}, o_t)$. These signals are combined into a compact control vector:
\begin{equation}
p_{mode}^t = W_1 H_{\le t} + W_2 o_t +W_3 I + b ,
\end{equation}
which directly parametrizes the reasoning mode logits. From this, the mode selection policy (cf. Equation~\ref{equ:ada_reason}) is given by:
\begin{equation}
\label{equ:pi_selection}
\pi_{\text{sel}}(m_t \mid h_t^{\text{rea}}, I) = \text{Softmax}(p_{mode}^t) .
\end{equation}



\subsubsection{Heuristic-to-RL Training}





Benefiting from the interpretable signals of the UAR Block, we do not require large-scale reasoning annotations. Instead, we propose a Heuristic-to-RL training mechanism, shown in Algorithm~\ref{alg:heuristic_rl}, that bootstraps policy learning with uncertainty-based heuristics. These priors provide a stable cold-start exploration, enabling the agent to discover useful reasoning patterns without exhaustive supervision. As training progresses, the heuristic influence is gradually annealed, allowing reinforcement learning to refine the reasoning trigger policy. This approach integrates the efficiency of heuristic guidance with the optimality of RL, leading to adaptive long-horizon reasoning strategies that generalize to novel environments.



\paragraph{Uncertainty-based Prior.}
In the cold-start phase, the RL policy has not yet learned meaningful mode selection. We therefore initialize training with an uncertainty-based prior. Intuitively, a higher entropy indicates a higher uncertainty, which requires stronger reasoning. We compute the scalar entropy score as the mean of past entropies, $H_{\text{score}} = \frac{1}{t} \sum_{k=1}^t H_k$, and map it into a soft prior distribution over $\lvert \mathcal{M} \rvert + 1$ reasoning modes (including the “no reasoning” option):
\begin{equation}
p_{\text{prior}} = \frac{\exp\big(-|H_{\text{score}} - \tau_m|/\sigma\big)}{\sum_{i=0}^{|\mathcal{M}|} \exp\big(-|H_{\text{score}} - \tau_i|/\sigma\big)} , \quad m = 0, \dots, |\mathcal{M}|
\end{equation}
where $\{\tau_0, \tau_1, \dots, \tau_{|\mathcal{M}|}\}$ are mode-specific entropy thresholds ($\tau_k=\tau_0+k\delta$), and $\sigma$ controls the smoothness of the prior.


\begin{wrapfigure}{r}{0.5\textwidth}
  \vspace{-32pt}
  \begin{minipage}{0.5\textwidth}
    \begin{algorithm}[H]
    \setlength{\baselineskip}{0.95\baselineskip} 
    \caption{Heuristic-to-RL}
    \label{alg:heuristic_rl}
    \begin{algorithmic}[1]
        \State Initialize navigation policy $\pi_{\text{nav}}$, reasoning selector $\pi_{\text{sel}}$, annealing schedule $\lambda_t$
        \For{each training episode $\tau$}
            \For{each step $t=1 \dots T$}
                \State Observe state $o_t$ and entropy $H_{\leq t}$
                \State Compute control vector $p_{\text{mode}}$
                \State Estimate Uncertainty prior $p_{\text{prior}}$
                \State Fuse prior and model (Eq.~\ref{equ:fuse_prior})
                \State Sample reasoning mode $m_t \sim p_{\text{final}}$
                \If{$m_t \neq \varnothing$}
                    \State Generate reasoning $r_t$
                \EndIf
                \State Select action $a_t \sim \pi_{\text{nav}}(a_t \mid o_t, r_{\leq t})$
                \State Execute $a_t$, observe next state $s_{t+1}$
                \State Compute extrinsic reward $r(s_t,a_t)$ and reasoning cost $c_{\text{rea}}(t)$
            \EndFor
            \State Compute task reward (Eq.~\ref{equ:return_task})
            \State Update policy (Eq.~\ref{equ:adaptive_reason})
        \EndFor
    \end{algorithmic}
    \end{algorithm}
  \end{minipage}
  \vspace{-14.5pt}
\end{wrapfigure}
\vspace{-3mm}
\paragraph{Heuristic-to-RL Transition.}
To gradually shift control from heuristic priors to learned RL policies, we fuse the prior distribution with the model prediction as:
\begin{equation}
\label{equ:fuse_prior}
p_{\text{final}}^t = \lambda_t \cdot p_{\text{prior}} + (1-\lambda_t) \cdot p_{\text{mode}},
\end{equation}
where $\lambda_t$ is annealed from $1$ to $0$ over training steps, allowing the RL policy $p_{\text{model}}$ to progressively take over from the uncertainty-based heuristic prior $p_{\text{prior}}$.  
Accordingly, Equation~\ref{equ:pi_selection} can be expressed as:
\begin{equation}
\pi_{\text{sel}}(m_t \mid h_t^{\text{rea}}, I) = \text{Softmax}(p_{\text{final}}^t).
\end{equation}
\paragraph{Reward Design.}
We first define the \textit{reasoning cost} as a normalized penalty based on the relative reasoning length:
\begin{equation}
c_{\text{rea}}(t) = \text{clip}\Bigg(\frac{L_t - L_{\text{shortest\_success}}}{L_{\text{window}}}, 0, 1 \Bigg)
\end{equation}
where $L_i$ is the reasoning length for the current step, $L_{\text{shortest\_success}}$ is the minimal generation length among success samples within the group, and $L_{\text{window}}$ is a constant penalty window.


For the navigation objective, we adopt the common extrinsic reward based on distance reduction, where the immediate reward is defined as
$r(s_t, a_t) = D_{\text{target}}(s_t) - D_{\text{target}}(s_{t+1}), ; t < T$,
with $D_{\text{target}}(s_t)$ denoting the geodesic distance from the current state $s_t$ to the target location $s_{\text{target}}$.

Finally, by combining extrinsic reward and reasoning cost, the overall task reward formulated in~\ref{equ:adaptive_reason} is defined as the discounted cumulative return:
\begin{equation}
\label{equ:return_task}
R_{\text{task}}(\tau) = \sum_{t=1}^{T} \beta^{\,t-1} \, \Big( r(s_t, a_t) - c_{\text{rea}}(t) \Big)
\vspace{-1mm}
\end{equation}
where $\beta \in (0,1]$ is the discount factor controlling the weight of future rewards. This formulation encourages the agent to navigate efficiently toward the goal while avoiding unnecessary reasoning overhead.

Overall, this Heuristic-to-RL scheme combines the efficiency of uncertainty-based priors with the optimality of RL, allowing the agent to progressively acquire adaptive reasoning strategies.

\vspace{-2mm}
\section{Experiments}
\vspace{-2mm}
We conduct experiments to answer the following questions:
(1) \textbf{Performance Gain}: How much does our proposed AdaNav improve over existing models on VLN-CE benchmarks and general spatial scene understanding tasks?
(2) \textbf{Reasoning Coordination}: What scheduling strategy has the UAR Block learned, and does it affect navigation efficiency?
(3) \textbf{Real-World Effectiveness}: How effective is AdaNav when deployed in real-world scenarios?
\vspace{-1mm}
\subsection{Performance Gain}
\vspace{-1mm}

\paragraph{Implementation details.}
\label{sec:detals_implementation}
\textbf{1. Base models.} AdaNav is designed to be general and can be integrated into existing VLN models with minimal modifications. To demonstrate its strong generalization ability, we adopt two SOTA open-source VLN models, \textsc{Navid}~\citep{zhang2024navid} and \textsc{Navila}~\citep{cheng2024navila}, as our base models.  
\textbf{2. Training setup.} Training is conducted on 4 NVIDIA RTX A100 GPUs. We construct the training set by randomly sampling 3,000 episodes from the training splits of both R2R~\citep{krantz2020beyond} and RxR~\citep{ku2020room}. For rollout collection during training, each episode is rolled out 5 turns, and the learning rate is set to $1\times10^{-6}$.
\textbf{3. Benchmarks.} To assess both navigation and spatial scene understanding, we evaluate on the val-unseen splits of R2R and RxR for navigation, and on the ScanQA validation set for scene understanding. Detailed settings are provided in Appendix~\ref{app:environment_metric}.

\paragraph{VLN-CE Benchmarks.}
We compare AdaNav with \textbf{recent million-scale closed-source models}, including \textsc{Navid-4D}~\citep{liu2025vid}, \textsc{Uni-Navid}~\citep{zhang2024uni}, and \textsc{MonoDream}~\citep{wang2025monodream}. As shown in Table~\ref{table:main_exp}, although closed-source models generally outperform open-source ones, AdaNav achieves substantial gains with only 6K training episodes, improving \textsc{Navid} and \textsc{Navila} by an average of 20\% on R2R and 14.6\% on RxR, respectively, and surpassing all closed-source baselines.

\paragraph{Cross-dataset Evaluation.}
As shown in Table~\ref{table:cross_data}, we test cross-dataset generalization by training AdaNav solely on 3K R2R samples and evaluating zero-shot on RxR Val-Unseen. AdaNav substantially improves base models, surpassing closed-source systems and demonstrating strong transferability.

\begin{table}[h] \fontsize{7}{8}\selectfont 
	\begin{center}
    \vspace{-2mm}
    \caption{Comparison with the state-of-the-art method on Val-Unseen split of R2R-CE and RxR-CE.}
    \vspace{-3mm}
	\label{table:main_exp}
	\renewcommand\tabcolsep{2.5pt}
\begin{tabular}{cccccccccccccc}
\toprule
 \multicolumn{1}{c|}{\multirow{2}{*}{Method}}              & \multicolumn{4}{c|}{Observation}                  & \multicolumn{4}{c|}{R2R-CE Val-Unseen}      & \multicolumn{4}{c|}{RxR-CE Val-Unseen}      & \multicolumn{1}{c}{\multirow{2}{*}{Training Data}} \\ \cline{2-13}
 \multicolumn{1}{c|}{}             & S.RGB & Pano. & Depth & \multicolumn{1}{c|}{Odo.} & NE↓ & OS↑ & SR↑ & \multicolumn{1}{c|}{SPL↑} & NE↓ & OS↑ & SR↑ & \multicolumn{1}{c|}{nDTW↑} & \multicolumn{1}{c}{}                               \\ \hline
\multicolumn{14}{c}{Open-Source}                                                                                                                                                                                                        \\ \hline
\multicolumn{1}{c|}{Seq2Seq}       &   \checkmark    &       &    \checkmark   & \multicolumn{1}{c|}{}     &  7.77 &37.0 &25.0 &\multicolumn{1}{c|}{22.0} &12.10 &13.9 &11.9 &\multicolumn{1}{c|}{30.8}           &-                                       \\
\multicolumn{1}{c|}{CMA}           &    \checkmark   &       &  \checkmark     & \multicolumn{1}{c|}{}         &7.37 &40.0 &32.0 & \multicolumn{1}{c|}{30.0} & -    &  -   &  -   & \multicolumn{1}{c|}{-}     &        -                                            \\
\multicolumn{1}{c|}{RGB-Seq2Seq}   &    \checkmark   &       &       & \multicolumn{1}{c|}{}     &10.10 &8.0 &0.0 &\multicolumn{1}{c|}{0.0}    &   -  &  -   &   -  & \multicolumn{1}{c|}{-}     &                 -                                   \\
\multicolumn{1}{c|}{RGB-CMA}       &     \checkmark  &       &       & \multicolumn{1}{c|}{}     & 9.55 &10.0 &5.0 &\multicolumn{1}{c|}{4.0}    &  -   & -    & -    & \multicolumn{1}{c|}{-}     &       -                                             \\
\multicolumn{1}{c|}{LAW}   &\checkmark       &       &  \checkmark     & \multicolumn{1}{c|}{\checkmark}     &6.83 &44.0& 35.0 & \multicolumn{1}{c|}{31.0} &10.90 &8.0 &8.0 & \multicolumn{1}{c|}{38.0}      &150K                                                \\
\multicolumn{1}{c|}{AO-Planner}    &       &   \checkmark    & \checkmark      & \multicolumn{1}{c|}{}     & 5.55 &59.0 &47.0 &\multicolumn{1}{c|}{33.0} &7.06 &43.3 &30.5 &\multicolumn{1}{c|}{50.1}    &       40K (Distill)                                             \\
\multicolumn{1}{c|}{NaVid}         &   \checkmark    &       &       & \multicolumn{1}{c|}{}     &    5.47 &49.0 &37.0 &\multicolumn{1}{c|}{35.0}    &  6.79   & 46.2    & 40.5    & \multicolumn{1}{c|}{52.2}     &  550K                                                  \\
\multicolumn{1}{c|}{NaVILA}        &    \checkmark   &       &       & \multicolumn{1}{c|}{}     &     5.22 &62.5& 54.0 &\multicolumn{1}{c|}{49.0} &6.77 &49.3 &44.0 &\multicolumn{1}{c|}{58.8}       &\textasciitilde 3000K                                        \\ \hline
\multicolumn{14}{c}{Close-Source}                                                                                                                                                                                                       \\ \hline
\multicolumn{1}{c|}{NaVid-4D}     &  \checkmark     & \checkmark      &       & \multicolumn{1}{c|}{}     &   5.99 &55.7 &43.8 & \multicolumn{1}{c|}{37.1}    & -    &   -  &  -   & \multicolumn{1}{c|}{-}     &      1840K                                              \\
\multicolumn{1}{c|}{Uni-NaVid}     &  \checkmark     &       &       & \multicolumn{1}{c|}{}         & 5.58  & 53.5  & 47.0  & \multicolumn{1}{c|}{42.7}  & 6.24  & 48.7  & 40.9  & \multicolumn{1}{c|}{-}     &  3600K                                                  \\
\multicolumn{1}{c|}{MonoDream}     &  \checkmark     &       &       & \multicolumn{1}{c|}{}     &    5.45 &61.5 &55.8 &\multicolumn{1}{c|}{49.1}   &6.38 &55.8 &49.4     & \multicolumn{1}{c|}{-}     &  1420K                                                  \\
\hline
\multicolumn{14}{c}{AdaNav}                                                                                                                                                                                                             \\ \hline
\multicolumn{1}{c|}{NaVid-AdaNav}  &  \checkmark     &       &       & \multicolumn{1}{c|}{}     & 5.39   &  57.89    & 47.7    & \multicolumn{1}{c|}{42.34}     &6.38     & 58.1    & 47.01    & \multicolumn{1}{c|}{56.8}     &     +6K                                               \\
\multicolumn{1}{c|}{NaVILA-AdaNav} &  \checkmark     &       &       & \multicolumn{1}{c|}{}     & 5.01    &  66.62   &  60.19   & \multicolumn{1}{c|}{50.0}     & 6.21    &60.51    &49.8    & \multicolumn{1}{c|}{62.2}     &     +6K                             \\ \bottomrule
\end{tabular}
 	\end{center}
 \vspace{-5mm}
\end{table}

\begin{table*}[h] 
\centering
\begin{minipage}{0.45\textwidth}
\centering
\fontsize{7}{8}\selectfont
\caption{Cross-dataset performance on the RxR-CE [30] ValUnseen split, without training on RxR-CE training set. * indicates our reproduction following the original papers. }
\label{table:cross_data}
\renewcommand\tabcolsep{3.5pt}
\begin{tabular}{ccccc}
\toprule
 \multicolumn{1}{c|}{\multirow{2}{*}{Method}}               & \multicolumn{4}{c}{RxR-CE Val-Unseen}   \\ \cline{2-5}
\multicolumn{1}{c|}{}              & NE↓ & OS↑ & SR↑ & SPL↑                  \\ \hline
\multicolumn{5}{c}{Open-Source}                                              \\ \hline
\multicolumn{1}{c|}{LAW}          &10.87 &21.0 &8.0 &8.0                       \\
\multicolumn{1}{c|}{CM2}           &8.98 &25.3 &14.4 &9.2                   \\
\multicolumn{1}{c|}{Seq2Seq}        &11.8& 5.02 &3.51 &3.43                       \\
\multicolumn{1}{c|}{CMA}          & 11.7& 10.7 &4.41 &2.47                      \\
\multicolumn{1}{c|}{NaVid*}          &8.57 &32.21 &21.3 &20.01                       \\
\multicolumn{1}{c|}{NaVILA*}      & 8.96 &43.35& 32.5& 26.82                       \\ \hline
\multicolumn{5}{c}{Close-Source}                                             \\ \hline
\multicolumn{1}{c|}{Uni-NaVid}     &8.08& 40.9 &29.5& 28.1                       \\
\multicolumn{1}{c|}{MonoDream}     & 8.57& 35.9& 25.1& 21.6                      \\ \hline
\multicolumn{5}{c}{AdaNav}                                                   \\ \hline
\multicolumn{1}{c|}{NaVid-AdaNav}  &  8.21   & 39.21    & 28.95    & 27.73 \\
\multicolumn{1}{c|}{NaVILA-AdaNav} &  8.25   & 48.65    &  38.82   & 31.21 \\ \bottomrule
\end{tabular}
\end{minipage}
\hfill
\begin{minipage}{0.51\textwidth}
\centering
\fontsize{7}{8}\selectfont
\caption{Evaluation of spatial scene understanding performance on the ScanQA Validation split. 
$^{*}$ and $^{\dagger}$ denote the use of 8 frames and 64 frames, respectively.}
\label{table:scanqa}
\renewcommand\tabcolsep{2pt}
\begin{tabular}{cccccc}
\toprule
\multicolumn{1}{c|}{\multirow{2}{*}{Method}}  & \multicolumn{5}{c}{ScanQA Validation}     \\ \cline{2-6} 
\multicolumn{1}{c|}{}                         & Bleu-4↑ & Rouge↑ & Cider↑ & Meteor↑ & EM↑ \\ \hline
\multicolumn{6}{c}{3D Large Multi-modal Models}                                           \\ \hline
\multicolumn{1}{c|}{3D-LLM}                   &7.2 &32.3 &59.2 &12.2 &20.4     \\
\multicolumn{1}{c|}{LL3DA}                    &13.5 &37.3 &76.8 &15.9         &  -   \\
\multicolumn{1}{c|}{Chat-3Dv2}                &14.0 &   - &87.6  &  -        &-     \\
\multicolumn{1}{c|}{Scene-LLM}                &12.0 &40.0 &80.0& 16.6 &27.2     \\
\multicolumn{1}{c|}{LEO}                      &13.2& 49.2& 101.4 &20.0 &24.5     \\ \hline
\multicolumn{6}{c}{2D Vision-Langauge-ActionModels}                                       \\ \hline
\multicolumn{1}{c|}{Uni-NaVid}                & -        &    45.74     &   94.72     &  19.24       & 28.01    \\
\multicolumn{1}{c|}{NaviLLM}                  & 12.0 &38.4& 75.9& 15.4& 23.0     \\
\multicolumn{1}{c|}{NaVILA*}          &14.8 &46.4 &95.1 &18.7 &27.0     \\
\multicolumn{1}{c|}{NaVILA$\dagger$}        &16.9 &49.3& 102.7 &20.1 &28.6     \\
\multicolumn{1}{c|}{NaVILA-AdaNav*}  &  15.33       &47.75        & 97.34       & 19.12         &27.27     \\
\multicolumn{1}{c|}{NaVILA-AdaNav$\dagger$} & 16.65         &    50.6 &  102.81      &21.25         & 29.57     \\ \bottomrule
\end{tabular}
\end{minipage}
 \vspace{-4mm}
\end{table*}

\begin{table}[h] \fontsize{7}{8}\selectfont 
	\begin{center}
    \caption{Comparing in three diverse real-world environments scenes.}
	\label{table:real_world}
    \vspace{-3mm}
	\renewcommand\tabcolsep{3pt}
\begin{tabular}{c|cccc|cccc|cccc}
\toprule
\multirow{3}{*}{Method} & \multicolumn{4}{c|}{Meeting Room}                            & \multicolumn{4}{c|}{Home}                                 & \multicolumn{4}{c}{Office}                              \\ \cline{2-13} 
                        & \multicolumn{2}{c}{Sample} & \multicolumn{2}{c|}{Complex} & \multicolumn{2}{c}{Sample} & \multicolumn{2}{c|}{Complex} & \multicolumn{2}{c}{Sample} & \multicolumn{2}{c}{Complex} \\ \cline{2-13} 
                        & NE↓          & SR↑          & NE↓            & SR↑           & NE↓           & SR↑          & NE↓            & SR↑           & NE↓           & SR↑          & NE↓           & SR↑           \\ \hline
Navid                   &   2.0           &   67.5          &    2.8           &   50.2           &      1.55        & 65.5             &   1.88               & 55.5              &   2.5           &  61.0             & 3.0              &  52.5            \\
Navid-AdaNav    &\baseline1.6	&\baseline78.5	&\baseline2.2	&\baseline60.5	&\baseline1.3	&\baseline78	&\baseline1.5	&\baseline75.5	&\baseline2	&\baseline70	&\baseline2.5	&\baseline66.5
           \\
Navila                  &  1.8            & 74  &    2.0           &       65       &         1.0     & 82.5          &    1.4           &82.5              &2.1              & 76.8             &2.2              &70              \\
Navila-AdaNav      & \baseline 1.0            &\baseline82.5   &\baseline 1.6              &\baseline73.5              &\baseline0.85              &\baseline95           &\baseline1.1               &\baseline88              &\baseline1.5              &\baseline87.6              &\baseline2.1              &\baseline75.5       \\ \bottomrule
\end{tabular}
 	\end{center}
 \vspace{-4mm}
\end{table}

\paragraph{Spatial Scene Understanding Benchmarks.}
As a general navigation agent, robust spatial scene understanding (e.g., object localization, referring, and spatial reasoning) is crucial. To verify whether AdaNav fine-tuning affects such capability, we evaluate on the ScanQA validation benchmark~\citep{azuma2022scanqa}, a widely used dataset for 3D question answering, as shown in Table~\ref{table:scanqa}. Results show that after Heuristic-to-RL training, AdaNav not only preserves its general scene understanding ability without using ScanQA training data, but also achieves slight improvements, indicating enhanced robustness and transferability.

\paragraph{Real-World Evaluation} 
To demonstrate the effectiveness of AdaNav in real-time settings, we conduct experiments in real-world environments using 25 sample or complex instructions. Each instruction requires the agent to complete 5--10 sequential landmark-following sub-tasks (e.g., ``After passing through the ticket gate, walk straight to the sofa, turn right, take the elevator, continue walking straight, pass through the door until reaching a supermarket, and finally stop at the counter''). Each instruction is executed three times across three types of environments: \textit{Meeting Room}, \textit{Home}, and \textit{Office}, following the protocol in prior works~\citep{cheng2024navila,zhang2024navid}. The results are summarized in Table~\ref{table:real_world}.

\vspace{-1mm}
\subsection{Analysis of UAR Block}
\vspace{-2mm}

To better understand how the UAR Block adapts over training, we conduct a systematic analysis using models trained with 2K, 4K, and 6K data. We focus on two aspects: (1) the frequency and distribution of reasoning invocations across different steps and reasoning modes, and (2) the tendency to trigger reasoning under different episode difficulty levels.
\vspace{-1mm}
\begin{figure}[t]
    \centering
    \begin{subfigure}[t]{0.49\linewidth}
        \centering
        \includegraphics[width=\linewidth]{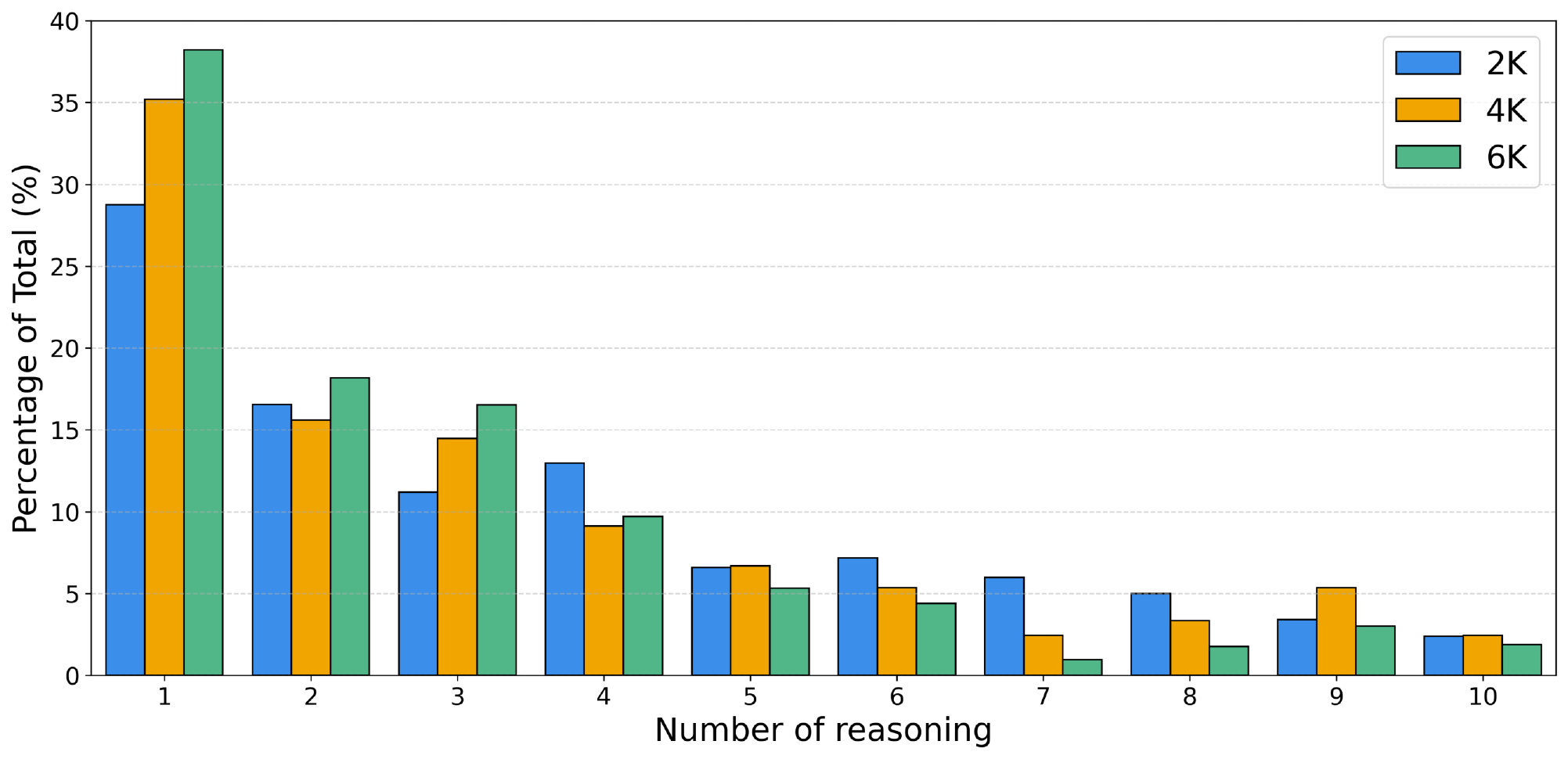}
        \caption{Distribution of reasoning steps per trajectory. As training data increases, the frequency of reasoning gradually decreases, with the model learning to invoke reasoning at more critical moments, thereby balancing efficiency and effectiveness.}
        \label{fig:reasoning_num}
    \end{subfigure}
    \hfill
    \begin{subfigure}[t]{0.49\linewidth}
        \centering
        \includegraphics[width=\linewidth]{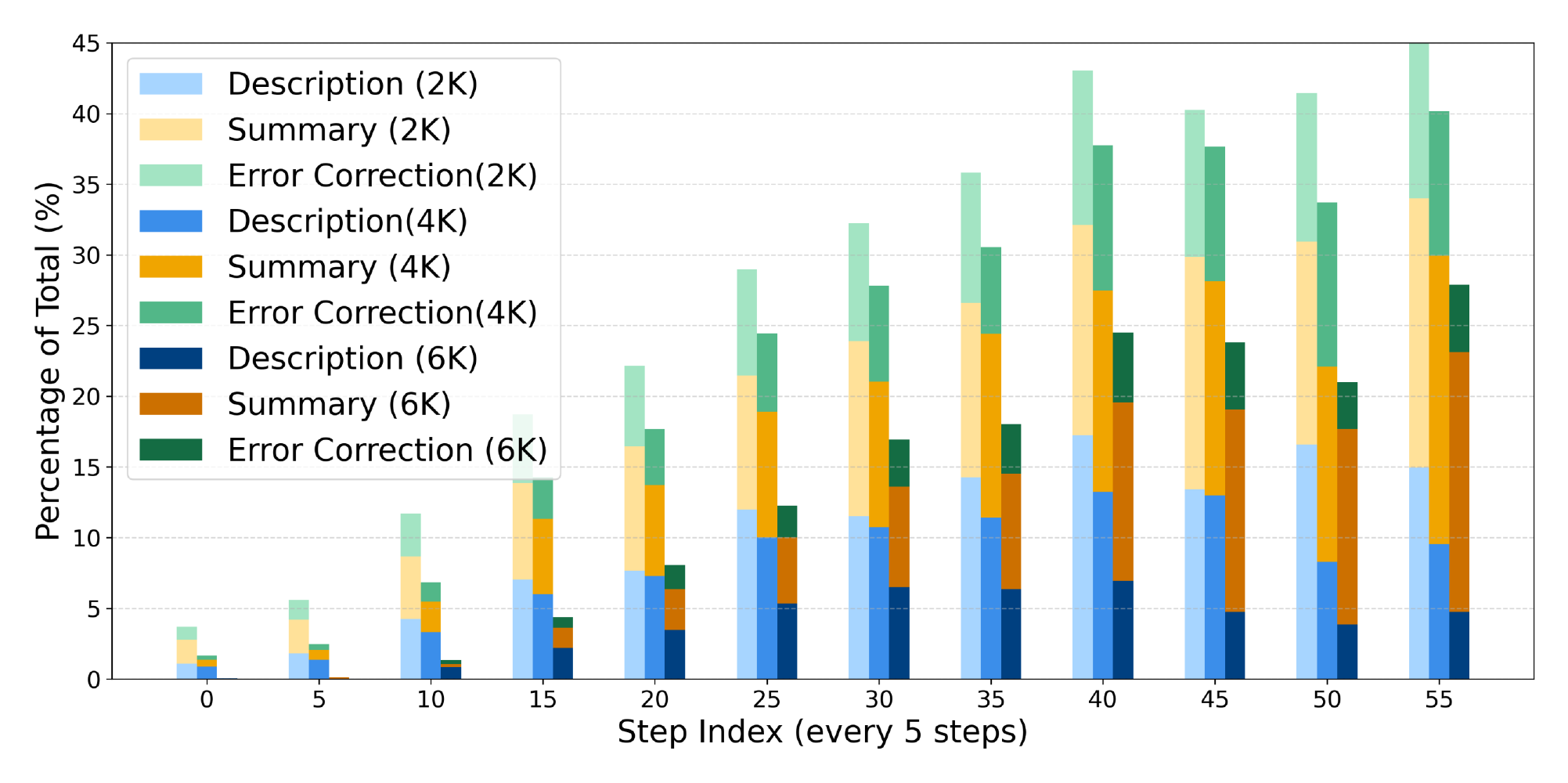}
        \caption{Number of reasoning invocations at each step, broken down by reasoning modes, across different training scales. As data increases, reasoning becomes more concentrated on high-uncertainty steps, with stronger preference for \textit{summary} mode at later stages.}
        \label{fig:reasoning_mode_step}
    \end{subfigure}
    \vspace{-2mm}
    \caption{Analysis of reasoning behaviors in AdaNav across training scales.}
    \label{fig:reasoning_analysis}
    \vspace{-5mm}
\end{figure}

\vspace{-2mm}
\paragraph{Frequency of Reasoning} Figure~\ref{fig:reasoning_num} shows the distribution of reasoning steps under different scales of training data, while Table~\ref{table:different_data_performance} reports the corresponding performance. As the training data increases, the model tends to reduce the frequency of reasoning, focusing more on triggering reasoning at critical moments, thereby balancing efficiency and effectiveness.

\vspace{-2mm}
\paragraph{Step-wise Reasoning Statistics.} 
Figure~\ref{fig:reasoning_mode_step} shows the number of reasoning invocations at each navigation step, broken down by mode (\textit{description}, \textit{summary}, \textit{error correction}) for the three training scales. We observe that as training data increases, the agent learns to concentrate reasoning on critical steps where uncertainty is higher, while reducing redundant reasoning in low-uncertainty steps. Additionally, the model increasingly favors \textit{summary} and \textit{error correction} modes at later steps, indicating adaptive mode selection based on task context.

\paragraph{Difficulty-conditioned Reasoning.}

\vspace{-3mm}
\begin{table*}[h] 
\centering
\begin{minipage}{0.80\textwidth}
\centering
\fontsize{8}{10}\selectfont
\caption{Proportion of reasoning triggers in hard episodes (success + failure = 100\%, excl. Step 0). Agents tend to invoke more reasoning in harder episodes.}
\vspace{-2mm}
\label{table:reasoning_success_fail}
\renewcommand\tabcolsep{3.5pt}
\begin{tabular}{c|cccccccccccc}
\toprule
Step & 0 & 5     & 10   & 15    & 20    & 25    & 30    & 35    & 40    & 45    & 50   & 55    \\ \hline
failure & 0 & 100.0 & 85.0 & 71.43 & 74.51 & 70.43 & 69.81 & 77.33 & 80.30 & 72.55 & 75.0 & 75.68 \\ \bottomrule
\end{tabular}
\end{minipage}
\hfill
\begin{minipage}{0.18\textwidth}
\centering
\fontsize{8}{10}\selectfont
\vspace{-2mm}
\caption{Performance on R2R.}
\label{table:different_data_performance}
\renewcommand\tabcolsep{2pt}
\begin{tabular}{c|ccc}
\toprule
Data & 2K & 4K & 6K \\ \hline
SR   & 44.8  & 46.5   &  47.7  \\ \bottomrule
\end{tabular}
\end{minipage}
 \vspace{-2mm}
\end{table*}

To disentangle how reasoning adapts to task difficulty, we first categorize each episode by its outcome (success vs. failure) under a baseline agent without reasoning coordination. We treat successful episodes as relatively easy and failed episodes as harder ones. We then re-run the model with the coordination layer enabled and analyze reasoning triggers across these two difficulty groups. 

As shown in Table~\ref{table:reasoning_success_fail}, for hard episodes that the base model fails to solve, reasoning is triggered significantly more frequently. This indicates that the UAR Block adaptively allocates reasoning capacity, focusing on challenging scenarios rather than applying reasoning uniformly across all episodes.

\paragraph{Conclusions.}
These analyses demonstrate that the UAR Block effectively learns both \emph{when} and \emph{which mode} to reason. As training progresses, reasoning becomes more temporally focused, mode-adaptive, and difficulty-aware, enabling the agent to improve navigation performance while minimizing redundant reasoning overhead.

\vspace{-2mm}
\section{Ablation}
\label{sec:ablation}
\vspace{-2mm}
To examine the robustness of AdaNav and assess whether its performance is overly dependent on specific hyperparameter choices, we conduct a series of ablation studies. Our analyses focus on three aspects: (1) component ablation, (2) sensitivity to key hyperparameters. More detailed ablation results are provided in Appendix \ref{app:ablation}.

\paragraph{Component Ablation.}
We use Navid as the base model and remove or replace major components to isolate their contributions. 
\textbf{(i) w/o UAR Block:} reasoning is invoked at a fixed step (5 step) interval or randomly, without adaptive control.
\textbf{(ii) w/o Heuristic Prior:} the agent relies purely on reinforcement learning from scratch without uncertainty-based heuristic. 
\textbf{(iii) w/o RL Fine-tuning:} reasoning triggers are guided only by heuristic signals without further policy refinement. 
Results show that removing either coordination or RL fine-tuning leads to significant performance degradation, confirming that both adaptive gating and learned refinement are essential.

\vspace{-2mm}
\begin{table*}[h] \fontsize{8}{9}\selectfont
    \caption{Ablation on hyperparameter sensitivity and component effectiveness on R2R-CE Val-Unseen. Here, * denotes fixed-interval (5 steps) triggering, and $\dagger$ denotes random triggering.}
    \centering
    \renewcommand\tabcolsep{3pt}
        \begin{subtable}[t]{0.33\textwidth}
        \centering
        \begin{tabular}{c|cccc}
        \hline
        Component & NE↓ & OS↑ & SR↑ & SPL↑ \\ \hline
        Navid   & 5.47 &49.0 &37.0 &35.0 \\
        w/o UAR*& 5.45 & 53.25 &40.12  & 38.83 \\
        w/o UAR$\dagger$ & 5.44 & 52.10 & 39.5 & 38.65 \\
        w/o RL  & 5.44 & 55.33  & 42.53 & 39.65 \\
        w/o HP  & 5.41 & 55.73 & 43.82 & 40.12 \\
        AdaNav  & \baseline 5.39   & \baseline 57.89    &\baseline 47.7    & \baseline 42.34   \\ \hline
        \end{tabular}
        \caption{Component ablation.}
        \label{table:component_ablation}
    \end{subtable}%
    \hfill
    \hspace{0.5mm}
    \begin{subtable}[t]{0.31\textwidth}
        \centering
        \begin{tabular}{cc|cccc}
        \hline
        \multicolumn{1}{c|}{$\tau_0$} & $\delta$ & NE↓ & OS↑ & SR↑ & SPL↑ \\ \hline
        \multicolumn{1}{c|}{\multirow{3}{*}{80\%}} &0.1  & 5.43 &57.72  & 48.82  &43.56  \\
        \multicolumn{1}{c|}{}     &0.2  &5.42  & 57.92 & 49.11 & 43.53 \\
        \multicolumn{1}{c|}{}     &0.3 &5.42  & 58.01 & 49.05 & 43.57 \\ \hline
        \multicolumn{1}{c|}{\multirow{3}{*}{85\%}} &0.1  & 5.40 &58.75  & 49.61  & 43.87  \\
        \multicolumn{1}{c|}{}     &0.2  & \baseline 5.39   &  \baseline57.89    &\baseline 47.7    & \baseline 42.34    \\
        \multicolumn{1}{c|}{}     &0.3  &5.39  & 58.81 & 49.53  & 43.85 \\ \hline
        \multicolumn{1}{c|}{\multirow{3}{*}{90\%}} &0.1  & 5.47 & 57.98 & 48.95 & 43.42  \\
        \multicolumn{1}{c|}{}     &0.2  & 5.48 & 57.80 & 48.83  & 43.34  \\
        \multicolumn{1}{c|}{}     &0.3  & 5.43 & 57.78 &48.85  & 43.35  \\ \hline
        \end{tabular}
        \caption{Effect of $(\tau_0,\delta)$.}
        \label{table:hyper_ablation1}
    \end{subtable}%
    \hfill
    \begin{subtable}[t]{0.33\textwidth}
        \centering
        \begin{tabular}{c|cccc}
        \hline
        $\sigma$ & NE↓ & OS↑ & SR↑ & SPL↑ \\ \hline
        0.05 & 5.43 &58.85  & 48.85& 43.55 \\
        0.10 & 5.40 &58.55  & 49.13 & 43.62\\
        0.15 & \baseline 5.39   &  \baseline57.89    &\baseline 47.7    & \baseline 42.34  \\
        0.20 & 5.41 & 58.73 & 49.55 &44.02  \\
        0.25 & 5.44 & 58.72 & 49.25 &43.88  \\
        0.30 & 5.48 & 58.13 & 48.72 & 43.92 \\ \hline
        \end{tabular}
        \caption{Effect of $\sigma$.}
        \label{table:hyper_ablation2}
    \end{subtable}
\vspace{-7mm}
\end{table*}

\paragraph{Hyperparameter Sensitivity.}
The key hyperparameters in our framework lie in the \textit{Heuristic-to-RL} stage, where we introduce mode-specific entropy thresholds: ($\tau_0$, $\delta$) that govern reasoning triggers prior, and a scaling factor $\sigma$.

As shown in Table~\ref{table:component_ablation}, a well reasoning prior significantly facilitates training. Specifically, $\tau_0$ is estimated from the base model by analyzing 1,000 validation episodes and selecting a percentile of the step-wise action entropy extrema. We experiment with percentiles at 80\%, 85\%, and 90\%, which define progressively stricter confidence thresholds. On top of this, $\delta$ incrementally shifts the thresholds for higher reasoning modes, thereby shaping the curriculum schedule. The corresponding results are summarized in Table~\ref{table:hyper_ablation1} and Table~\ref{table:hyper_ablation2}.

\vspace{-2mm}
\section{Conclusion}
\vspace{-2mm}
In this work, we tackled the long-standing challenges of consistent temporal grounding and robust perception–action mapping in Vision-Language Navigation. We proposed AdaNav, an uncertainty-based adaptive reasoning framework that integrates the UAR Block with a Heuristic-to-RL training mechanism. This design enables agents to invoke reasoning adaptively, guided first by interpretable heuristic priors and then refined through reinforcement learning, without relying on costly labeled reasoning data. Extensive experiments show that AdaNav delivers substantial improvements: surpassing million-scale closed-source models with only 6K samples, generalizing effectively across R2R and RxR, and demonstrating strong robustness in real-world deployment. Moreover, AdaNav reduces reasoning frequency while making it more difficulty-aware and mode-adaptive, striking a balance between efficiency and effectiveness. AdaNav provides a principled and practical step toward scalable, adaptive reasoning in embodied agents.



\bibliography{iclr2026_conference}
\bibliographystyle{iclr2026_conference}

\appendix

\section{Implementation Details}
\label{app:environment_metric}


\paragraph{Real-World Evaluation}
\label{app:real_world}
\begin{wrapfigure}{r}{0.45\columnwidth}  
  \includegraphics[width=0.4\columnwidth]{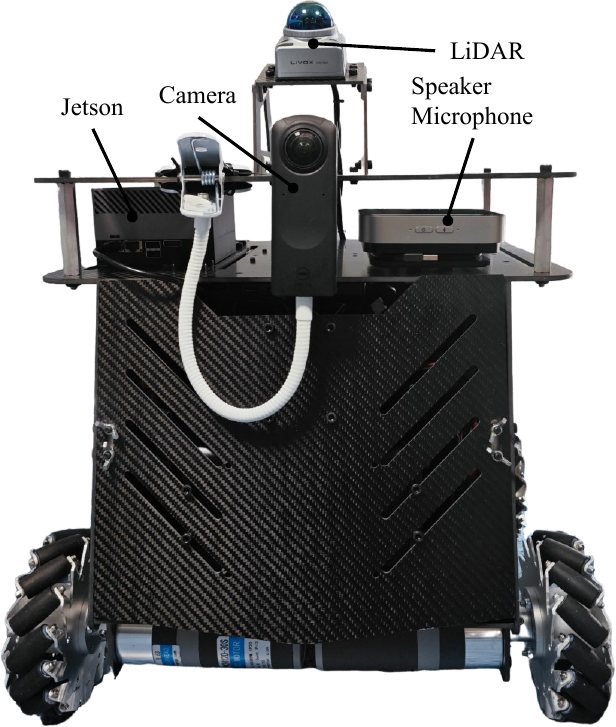}
  \caption{Hardware setup of the mobile robot platform used for real-world evaluation.}
  \vspace{-5mm}
  \label{app:robot_setup}
\end{wrapfigure}
We provide a detailed description of our real-robot platform in Figure~\ref{app:robot_setup}. The mobile robot is equipped with basic locomotion capabilities and augmented with a camera, microphone, speaker, and LiDAR sensors for user interaction and environment perception. Notably, our method relies only on RGB images and does not require LiDAR input. The system is powered by a Jetson AGX Orin~\citep{nvidiaJetsonNano} running Ubuntu 24.04 with ROS2 Jazzy. In addition, the platform integrates a 19V power regulator and a 110V/220V inverter (both rated at 500W+) to support the compute and sensor modules.

Building on this platform, we design a pipeline for vision-and-language navigation with AdaNav. We experiment with AdaNav on two base models, \textsc{Navid}~\citep{zhang2024navid} and \textsc{Navila}~\citep{cheng2024navila}. In deployment, AdaNav runs on a server with a Jetson that receives compressed images from the robot and sends back high-level commands. The robot then executes these commands (e.g., “Turn right” or “Move forward”) through its locomotion system. During navigation, the robot continuously monitors its motion to ensure that rotations and forward movements remain aligned with the issued commands.
\begin{figure}[h]
\centering
\includegraphics[width=1.0\textwidth]{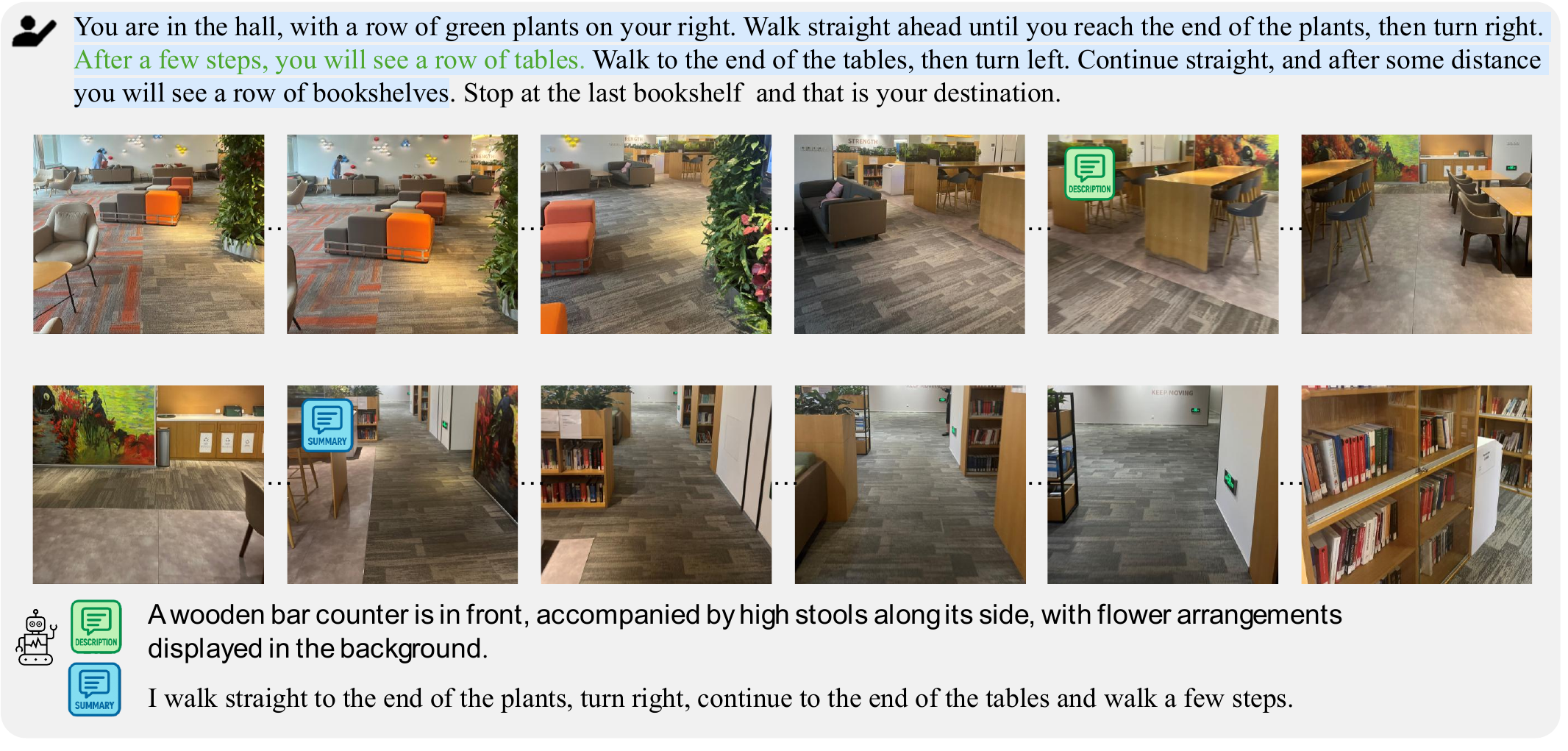}
\caption{Visualization of adaptive reasoning navigation in real-world scene.}
\label{fig:visual_reason}
\vspace{-5mm}
\end{figure}

\section{Detailed Ablation Analysis for Robustness and Sensitivity}
\label{app:ablation}
To further analyze our method, we conduct ablations on different reasoning modes. As shown in Figure~\ref{fig:reasoning_mode_step}, different modes contribute at different stages of navigation. This raises the question: \emph{how much would performance degrade if only a single reasoning mode were available?} To investigate, we construct variants where only one reasoning mode is enabled (i.e., description, summary, or error correction). In addition, we add a control setting, where the model is simply instructed with a generic prompt such as: \textit{"At this point, perform some analysis based on the past trajectory."} For all these variants, we apply UAR Block together with the heuristic-to-RL training mechanism, and report the results as follows.

\begin{table}[h]\fontsize{8}{9}
    \centering
    \caption{Ablation results under different \textit{reasoning modes}, where \mone: Description only, \mtwo: Summary only, \mthree: Error correction only, and \mfour: Generic prompt reasoning. All variants are trained with UAR Block and the heuristic-to-RL mechanism.}
    \label{app:mode_ablation}
    	\renewcommand\tabcolsep{3pt}
    \begin{tabular}{ccccc}
\toprule
 \multicolumn{1}{c|}{\multirow{2}{*}{Method}}               & \multicolumn{4}{c}{R2R-CE Val-Unseen}   \\ \cline{2-5}
\multicolumn{1}{c|}{}              & NE↓ & OS↑ & SR↑ & SPL↑                  \\ \hline
\multicolumn{1}{c|}{AdaNav}   & \baseline 5.39   & \baseline 57.89    &\baseline 47.7    & \baseline 42.34   \\
\multicolumn{1}{c|}{AdaNav \mone}    &5.41 &55.34 &44.52 &40.88                              \\
\multicolumn{1}{c|}{AdaNav \mtwo}  &  5.42 &56.53 &46.01 &41.37\\
\multicolumn{1}{c|}{AdaNav \mthree}  &  5.40 &56.12 &45.77 &41.67\\
\multicolumn{1}{c|}{AdaNav \mfour}  & 5.40 &55.23 &45.62 &42.01 \\ 
\bottomrule
\end{tabular}

\end{table}

\section{Visualization Results on Adaptive Reasoning Navigation}
\label{app:visual section appendix}

\begin{figure}[h]
\centering
\begin{minipage}{\textwidth}
    \centering
    \includegraphics[width=0.95\textwidth]{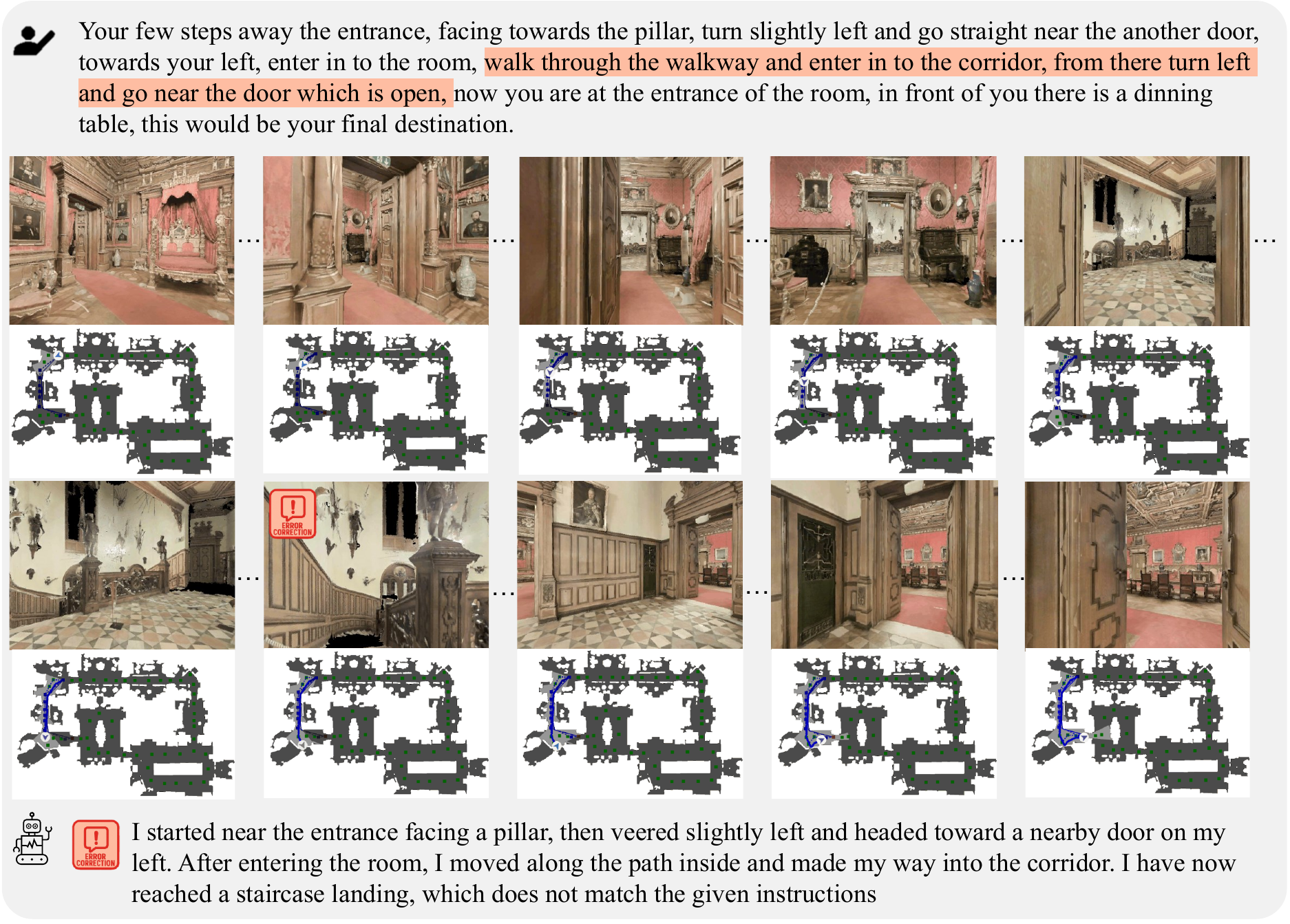}
\end{minipage}

\vspace{2mm} 

\begin{minipage}{\textwidth}
    \centering
    \includegraphics[width=0.95\textwidth]{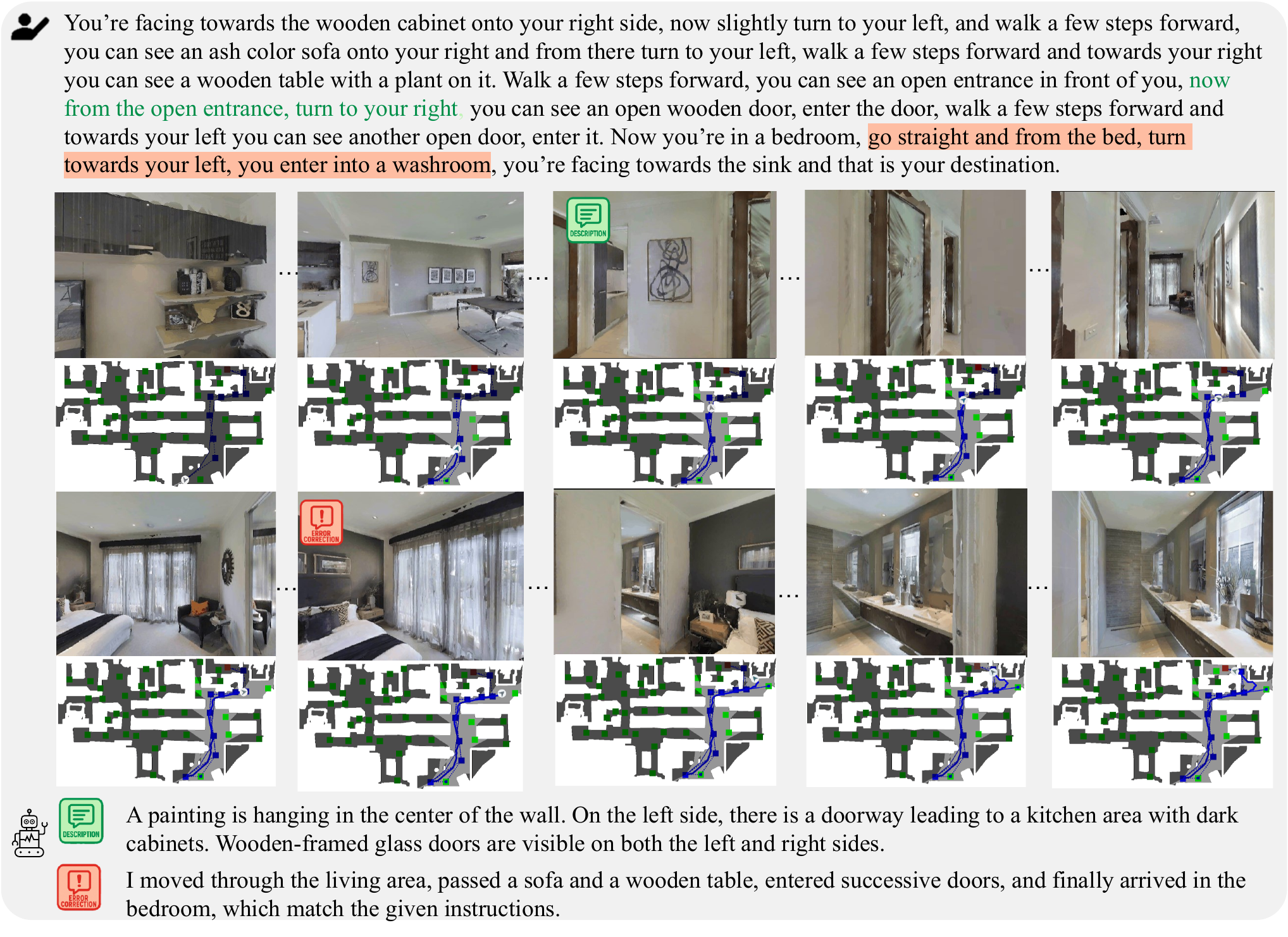}
\end{minipage}

\caption{Visualizations of adaptive reasoning navigation, where AdaNav autonomously invokes reasoning at high-uncertainty points to better align the trajectory with the instruction.}
\label{app:visual_reason}
\vspace{-5mm}
\end{figure}

To better understand how AdaNav adaptively allocates reasoning during navigation, we visualize example trajectories in Figure~\ref{app:visual_reason}. The figure illustrates both the agent's path and the steps where reasoning is invoked. As shown, AdaNav selectively triggers reasoning at critical or challenging moments, while skipping unnecessary steps in simpler regions of the environment. This demonstrates that the Uncertainty-Adaptive Reasoning (UAR) Block effectively guides the agent to balance efficiency and accuracy. 

Figure~\ref{app:visual_reason} also highlights that reasoning is concentrated on hard trajectories: compared to the first, simpler scenario, the second, more complex instruction involves more reasoning steps. This observation is consistent with our quantitative analysis in Table~\ref{table:reasoning_success_fail}. These visualizations confirm that the agent’s reasoning behavior is both difficulty-aware and mode-adaptive, providing interpretability and insight into its decision-making process.

\end{document}